\documentclass[10pt,twocolumn,letterpaper]{article}

\usepackage{cvpr}              
\usepackage{multirow}
\usepackage{booktabs}
\usepackage{amssymb}
  
\definecolor{cvprblue}{rgb}{0.21,0.49,0.74}
\usepackage[pagebackref,breaklinks,colorlinks,allcolors=cvprblue]{hyperref}
\usepackage{colortbl}

\def\paperID{1663} 
\def\confName{CVPR}
\def\confYear{2025}

\title{SAM2-LOVE: Segment Anything Model 2 in Language-aided \\Audio-Visual Scenes}

\author{Yuji Wang$^1$\footnotemark[1]~, Haoran Xu$^2$\footnotemark[1]~, Yong Liu$^1$, Jiaze Li$^2$, Yansong Tang$^1$\footnotemark[2]~\\
$^1$Tsinghua Shenzhen International Graduate School, Tsinghua University, $^2$ZJU\\
{\tt\small yuji-wan24@mails.tsinghua.edu.cn,  \textsuperscript{†}tang.yansong@sz.tsinghua.edu.cn}
}

\begin{document}
\maketitle
\footnotetext[1]{Equal contribution}
\footnotetext[2]{Corresponding author}
\begin{abstract}
Reference Audio-Visual Segmentation (Ref-AVS) aims to provide a pixel-wise scene understanding in Language-aided Audio-Visual Scenes (LAVS). This task requires the model to continuously segment objects referred to by text and audio from a video. Previous dual-modality methods always fail due to the lack of a third modality and the existing triple-modality method struggles with spatio-temporal consistency, leading to the target shift of different frames. In this work, we introduce a novel framework, termed \textbf{SAM2-LOVE}, which integrates textual, audio, and visual representations into a learnable token to prompt and align SAM2 for achieving Ref-AVS in the LAVS. Technically, our approach includes a multimodal fusion module aimed at improving multimodal understanding of SAM2, as well as token propagation and accumulation strategies designed to enhance spatio-temporal consistency without forgetting historical information. We conducted extensive experiments to demonstrate that SAM2-LOVE outperforms the SOTA by 8.5\% in $\mathcal{J\&F}$ on the Ref-AVS benchmark and showcase the simplicity and effectiveness of the components. Our code will be available \href{https://github.com/VoyageWang/SAM2LOVE}{here}.




\end{abstract}

\section{Introduction}
\label{sec:intro}

\begin{figure}[t]
	\centering	\includegraphics[width=0.5\textwidth]{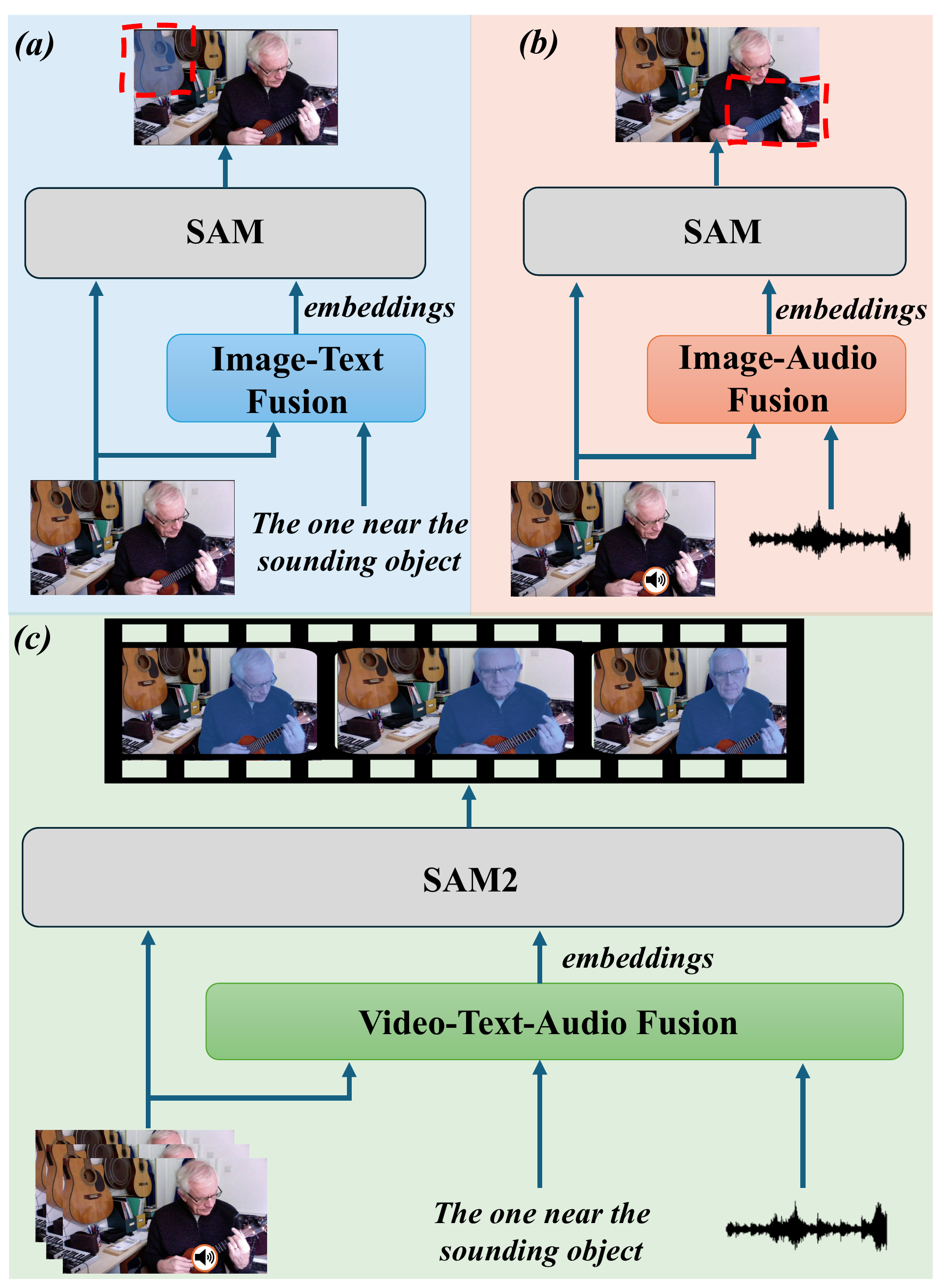}
	\caption{
    The comparison with dual-modality SAM pipelines. (a) The text-visual SAM struggles to distinguish the sounding object in the silent scenes. (b) The audio-visual SAM fails to understand the dynamic control from the text input. (c) Our model fuses three modalities' information and expands to the video due to SAM2.}
	\label{example}
\end{figure} 
\begin{figure*}[ht]
	\centering	\includegraphics[width=0.95\textwidth]{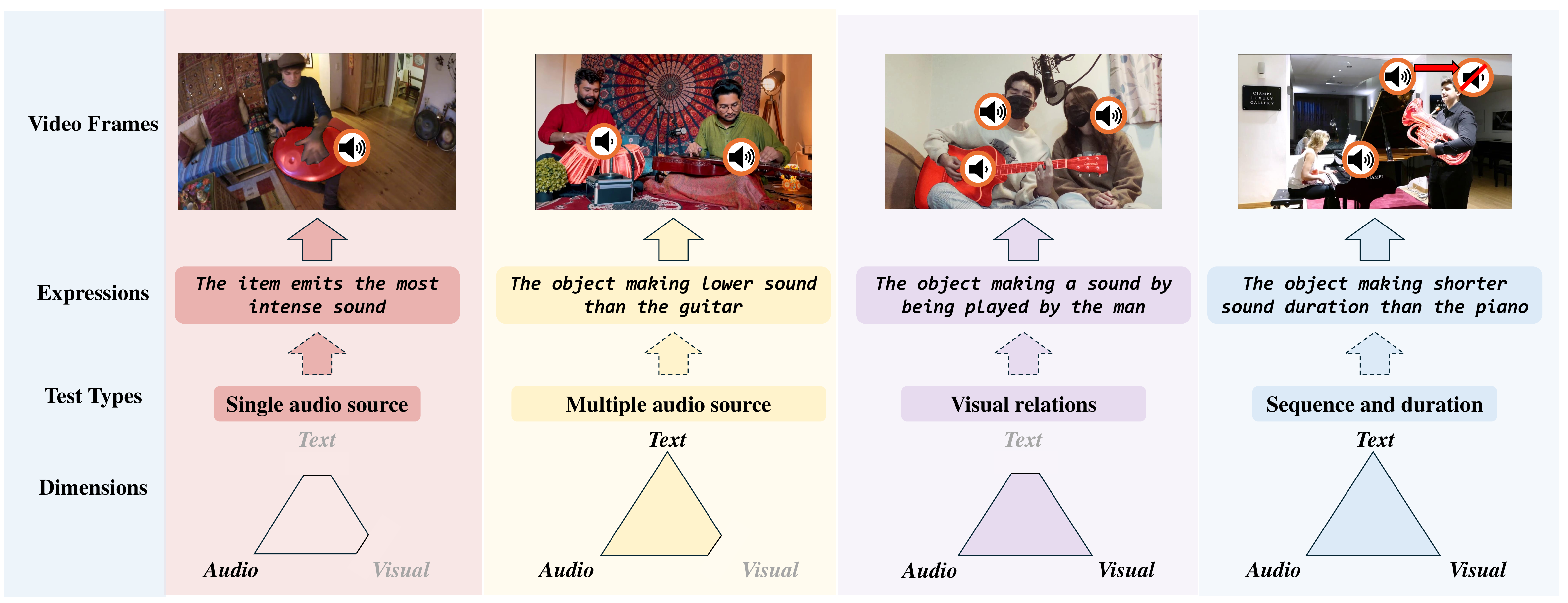}
	\caption{Illustration of the Ref-AVS task under varying modality cues. The expression can be treated as a kind of control signal to condition the referred object on audio. Different expressions can test the models' pixel-wise scene understanding capabilities in the LAVS with different dimensions. The absence angle of a triangle means that modality is not the main test dimension.}
	\label{reason}
\end{figure*}

In the process of human perception of the world, visual signals play a primary role, often accompanied by other multimodal signals such as audio and text. These additional multimodal cues are not only essential to locating and separating objects and areas of interest but also to understanding the changing behaviors of the objects. The dynamic scene that integrates visual elements with diverse multimodal cues including audio and text is regarded as the Language-aided Audio-Visual Scene (LAVS). To mimic the effect from humans, prior works such as Music-AVQA~\cite{Li_2022_CVPR} and AVQA~\cite{yang2022avqa} have primarily focused on question-answering tasks. Nevertheless, these approaches lack a fine-grained understanding of LVAS at the pixel level. As shown in Figure \ref{reason}, Reference Audio-Visual Segmentation (Ref-AVS) task \cite{wang2024refavs} makes a good example to improve the pixel-wise scene understanding of the model in the LAVS. The previous duality-referring segmentation works struggle with the Ref-AVS task where the interested object can only be located under triple modalities, shown in Figure \ref{example}. In neither Figure \ref{example} (a) text-visual EVF-SAM ~\cite{zhang2024evf} nor Figure \ref{example} (b) audio-visual GAVS~\cite{wang2024prompting}, the truly referred target is accurately segmented due to the absence of additional help from the other third modality. Additionally, they are constrained in the image level, which can not be directly adapted to the dynamic nature of video content. Therefore, methods for effectively fusing information from the three modalities for segmentation have garnered widespread attention.

Prior work, EEMC \cite{wang2024refavs}, simultaneously models three modalities' representations to achieve excellent results on the Ref-AVS benchmark. However, it showcases the relatively insufficient spatio-temporal consistency. Specifically, despite the existence of memory caches, the model fails to continuously track the target’s position and shape, leading to a shift in the segmented area over time. Recently, SAM2~\cite{ravi2024sam2} has been proposed to generalize the segmenting anything capabilities of the image to the video, which demonstrates a powerful video tracking capability in inter-frame consistency. SAM2 follows the previous promptable visual segmentation of SAM~\cite{sam} that requires a strong manual annotation to the interested target of the starting frame (not only limited to the first frame) and then propagates to the whole video based on the crucial frame. Nevertheless, the lack of dependency on other modalities (text or audio) constraints its further application in the LAVS.

Thus, in this paper, we propose a novel framework, namely \textbf{SAM2-LOVE}, which means \textbf{S}egment \textbf{A}nything \textbf{M}odel \textbf{2} in \textbf{L}anguage-aided audi\textbf{O}-\textbf{V}isual sc\textbf{E}nes. To enhance SAM2's understanding of multimodal information, we design the multimodal fusion module to compress all the modalities' information into a learnable token that aligns the SAM2 for segmenting the referring target. Additionally, we incorporate token propagation and accumulation strategies to strengthen spatio-temporal understanding of the video while avoiding forgetting historical frames. Our main contributions can be summarized as follows:
\begin{itemize}
\item We propose a novel framework, \textbf{SAM2-LOVE} that firstly leverages SAM2 to achieve pixel-wise understanding in the LAVS by designing a multimodal fusion module.

\item We develop creative token propagation and accumulation strategies to improve spatio-temporal comprehension of the promtable token.

\item Extensive experiments on Ref-AVS dataset demonstrate the superiority of our method, with ablation studies highlighting the simplicity and effectiveness of its modules.
\end{itemize}

\section{Related Works}
\label{sec:related}
\noindent\textbf{Segmentation via Language or Audio.}
Unlike semantic segmentation \cite{wang2023efficient,wang2024convolution,liu2024open,bai2024self}, visual segmentation via language starts with referring expression segmentation task~\cite{onlinerefer,kazemzadeh2014referitgame,wu2022language,nagaraja2016modeling,yang2022lavt,yang2024language, wang2025iterprime,liu2024universal} which performs target object segmentation based on given natural language expressions. X-Decoder~\cite{zou2023generalized} bridges vision and language, unifying multiple tasks within a single model. Moreover, GRES~\cite{GRES} defines a new task called generalized reference expression segmentation. With the development of multimodal large language models (MLLMs) such as LLaVA~\cite{liu2024visual,liu2024improved}, LISA~\cite{lai2024lisa} proposes the concept of reasoning segmentation with combination of pixel-level understanding and MLLMs. GSVA~\cite{Xia_2024_CVPR} and PixelLM~\cite{Ren_2024_CVPR} propose multi-target and null-target reasoning segmentation. On the other hand, audio-visual segmentation (AVS) is also constantly improving. Audio-Visual Segmentation~\cite{zhou2022avs,zhou2023avss} proposes the AVS problem and AVSBench dataset. Numerous subsequent methods~\cite{chen2024unraveling,gao2024avsegformer,sun2024auto} have been developed and refined based on the proposed benchmark.

\noindent\textbf{Segment Anything Model.} SAM~\cite{sam} is an interactive segmentation model based on various types of prompts (points, boxes, coarse
masks). It shows significant generalization ability in open-world segmentation. Several works strive to reduce the massive computation cost of SAM, such as EfficientSAM~\cite{efficientsam}, FastSAM~\cite{zhao2023fast}, and MobileSAM~\cite{mobile_sam}. The advent of SAM2~\cite{ravi2024sam2} extends SAM’s functionality to include video segmentation. While SAM and SAM2 perform effectively in visual segmentation tasks using box, point, or mask prompts, they currently do not possess language understanding capabilities, rendering the direct use of text prompts for referring segmentation infeasible. EVF-SAM~\cite{zhang2024evf} extends SAM's capabilities with text-prompted segmentation, while GAVS~\cite{wang2024prompting} equips the SAM with image-audio fusion to achieve AVS.

\noindent\textbf{Language-aided Audio-Visual Scene.} There is presently a scarcity of publicly available tasks that provide datasets for audio-visual scene comprehension with language-based support. The complexity of language-aided audio-visual scene understanding arises from its reliance on three modalities (vision, audio, and text) necessitating massive matched data and their corresponding labels, posing significant challenges for data collection and model design. Music-AVQA~\cite{Li_2022_CVPR} and AVQA~\cite{yang2022avqa} are two of the classic audio-visual question-answering (QA) datasets. However, the above work focuses on multimodal QA and spatio-temporal reasoning of audio-visual scenes involving various relationships between objects or activities. Ref-AVS~\cite{wang2024refavs,wang2024prompting} begins pixel-level understanding of dynamic audio-visual scenes.

In summary, no methods exist for integrating the foundational visual segmentation model SAM2 into language-based audio-visual scenes. Therefore, to achieve pixel-level understanding in the LAVS, \textbf{we introduce SAM2-LOVE}.


\section{Method}

\begin{figure*}[t]
	\centering	\includegraphics[width=1\textwidth]{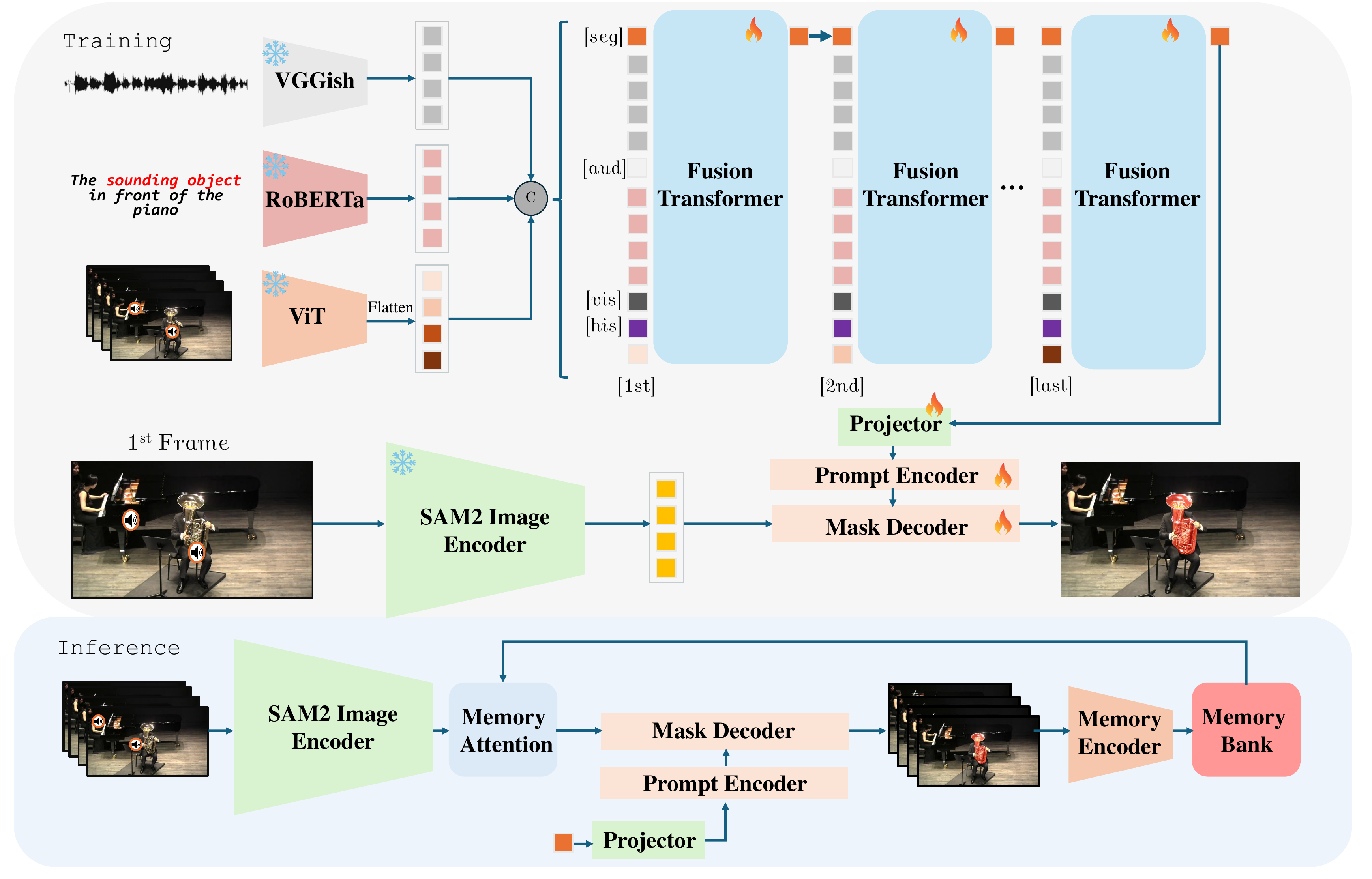}
	\caption{  The pipline of SAM2-LOVE. We harness a fusion transformer to compress the tripe modality representations into a learnable token $[seg]$. For the input, $[aud]$ and $[vis]$ are added to indicate their corresponding modality sequences, and $[his]$ is to provide additional information about former frames. In the training stage, we use the first frame to align the SAM2 with the compressed and fused representation of $[seg]$. During inference, we harness the strong zero-shot capabilities of SAM2 to track the referred target at the pixel level.}
	\label{pipleline}
\end{figure*} 

\subsection{Preliminary of SAM2}
SAM2 can be treated as a generalization of SAM to segment anything in images and videos. It also supports visual prompts (point, box, and mask) to indicate the objects to be segmented in an individual frame (image as a single frame). Then, it generalizes across the whole video based on the prompted frame to achieve the video segmentation. 

As shown in Figure \ref{pipleline} (Inference), this model begins with a hierarchical image encoder that encodes each frame into multilevel embeddings. Unlike SAM which directly feeds these features to the mask decoder for predictions, SAM2 incorporates a series of transformer blocks (i.e. memory attention) to condition the current image feature on memories of previous and prompted frames stored in a memory bank that is maintained by a FIFO (First In First Out) queue. Each memory entry is composed of two kinds of memories: (1) spatial feature maps that are generated through an element-wise summation of the downsampled and encoded mask features from the memory encoder and the unconditioned frame embedding from the image encoder, and (2) the object pointers that are obtained from the output tokens within the mask decoder contain high-level semantic information of the interested target. Therefore, the memory attention operation in the transformer makes the current frame's feature cross-attend both spatial fine-grained memory and abstract temporal object pointers. Second, the mask decoder of SAM2 largely follows the design of SAM, which is a ``two-way" transformer that takes the prompt embeddings from the prompt encoder and frame embeddings as inputs. The difference lies in the additional occlusion score that aims to judge the object's presence in the current frame.

Thanks to the strong and generalized capability of SAM2 across images and videos, we can first compress all the information contained in three modalities into a single token and then use it to prompt SAM2 on a single frame. Finally, starting from this crucial frame, we can propagate through video to obtain the desired masks of all frames.

\subsection{SAM2-LOVE}
The overall pipeline of our model is shown in Figure \ref{pipleline}, which integrates multimodal information in the LAVS into a representative token and uses it to provide a visual clue for SAM2 to segment the reference targets. 

\textbf{Multimodal Representations.} Following the previous work \cite{wang2024refavs}, for each modality, we use three popular encoders to separately extract their representations, which were not aligned before in their pretraining stages. VGGish~\cite{gemmeke2017audio, hershey2017cnn} encodes an audio clip into audio embeddings ${F_A} \in {R^{N \times {d_A}}}$, where $N$ denotes the duration of the audio in seconds and corresponds to the total number of frames in the video. The sampled $N$ frames are encoded into ${F_V} \in {R^{ N\times {d_V} \times h \times w}}$ by a ViT \cite{dosovitskiy2020image} pre-trained on ImageNet. For the text expressions, we utilize  DistilRoBERTa \cite{sanh2019distilbert,liu2019roberta} to obtain the textual representations ${F_T} \in {R^{P \times {d_T}}}$, where $P$ is the number of textual tokens. To unify and align all embeddings into the same semantic space, we use three MLPs to project them into a shared dimension respectively, shown in Equation (\ref{eq1}).
\begin{equation}
    \label{eq1}
      {{\widehat{F}_Z}} = Linear({\mathop{\rm ReLU}(Linear({F_Z}))),Z \in \{ A,V,T\}}
\end{equation}

\textbf{Multimodal Fusion.} As shown in Figure \ref{pipleline}, a multimodal fusion module is proposed to compress all three modalities' information into a single token and then leverage its embedding to represent the fused features. We define this token as a learnable segmentation token, $[seg]$, designed to prompt SAM2 as a clue to retrieve the targeted object. The fusion module is a naive transformer encoder architecture with $L$ stacked bi-directional transformer block that contains a multi-head self-attention (MSA), Layernorm (LN), and MLP, formulated below:
\begin{equation}
    \begin{array}{l}
z{'_l} = MSA(LN({z_{l - 1}})) + {z_{l - 1}},\quad l = 1...L\\
{z_l} = MLP(LN(z{'_l})) + z{'_l}, \quad l = 1...L
\end{array}
\end{equation}
where $z_{l-1}$ is the input sequence of $l$-th layer.

\begin{figure*}[t]
	\centering	\includegraphics[width=\textwidth]{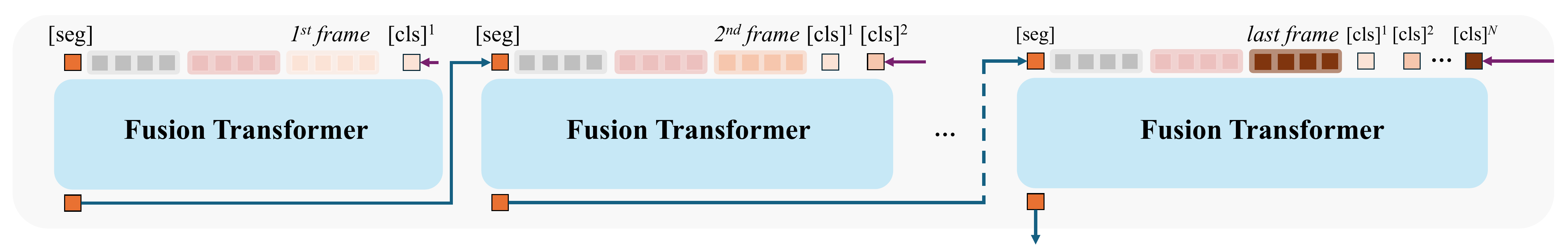}
	\caption{The token propagation and token accumulation strategy in the fusion transformer. For the token propagation, the output $[seg]$ of the current step serves as the input for the next step, which interacts with different frame features but the same audio and text features. To offer additional historical information, the token accumulation strategy appends the global representations of all the previous frames to update the maintained history token $[his]$.}
	\label{token_behaviour}
\end{figure*} 

\textbf{Token Propagation.} The behavior of the $[seg]$ token is a self-propagation process within the whole video, which adaptively captures the inner-frame spatial features and the strong continuity among consecutive frames to achieve spatio-temporal consistency. To be more specific, after obtaining the three projected multimodal representations $\widehat{F}_A$, $\widehat{F}_T$, and $\widehat{F}_V$, we first prepend the $[seg]$ token to the head of the sequence of modalities’ embeddings to obtain the multimodal sequence $F_M^i$ for the $i$-th frame, shown as follows:
\begin{equation}
    \label{eq2}
    F_M^i = Concat([ [seg] ;{\widehat{F}_A};[aud];{\widehat{F}_T};[vis];\widehat{F}_V^i]),
\end{equation}
where $Concat$ denotes the concatenation operation in the feature dimension axis, two fixed tokens, $[aud]$ and $[vis]$ indicate the end and beginning of their corresponding sequences, respectively. This concatenated sequence will be fed into the fusion transformer for updating the status of the $[seg]$ token, allowing the token to capture the visual information of the current frame along with the integral audio and text representations. Since the transformer is the bi-directional encoder that enables all the tokens to interact with each other, we can harness the single token to serve as the multimodal representation of the whole sequence. We take the first element from the output of the fusion transformer $Q_i = Transformer(F_M^i)$, as the updated $[seg]$ token $[seg] = Q_i^0$, shown in Figure \ref{token_behaviour}. Therefore, the token not only contains the current frame information but also the compressed information of the previous frames. Our proposed token propagation strategy allows it to model the video from both the spatial and temporal aspects and aggregate the information within the text and audio.

\textbf{Token Accumulation.} Although the token propagation strategy can effectively obtain the multimodal characteristics and be used to prompt the SAM2, it still faces the issue that the information of the first several frames can be diluted and forgotten with the increasing propagation times. To address the issue, we propose the $[cls]$ token accumulation strategy, in which we keep a history of token sequences, $[his]$ to store the global and abstract representation of historical frames, shown in Figure \ref{token_behaviour}. It is achieved that when propagating to the $i$-th frame, we provide additional global information of all the previous frames by appending and concatenating the $[his]$ tokens into $F_M^i$. The updation of the history token $[his]$ can be formulated by Equation (\ref{eq3}).
\begin{equation}
    \label{eq3}
    [his]^i =Concat([[cls]^0;[cls]^1;...;[cls]^i])
\end{equation}
where $[cls]^i$ is the global representation of $i$-th frame from ViT. These historical tokens can be referred to by the $[seg]$ of the current stage, promising the information of all the previous frames is not lost.

\textbf{Unified Perspective of Token Behaviors.} We interpret the two token operations described above from a unified knowledge transfer perspective. Token propagation of the $[seg]$ token is a direct flow, which expects the knowledge from different frames to be gradually connected through the propagation process. This is the \textbf{forward} knowledge transfer. In contrast, token accumulation is the replay of $[his]$ tokens, a reversed flow to prevent the $[seg]$ token from forgetting the knowledge of historical frames as it propagates forward. This is the \textbf{backward} knowledge transfer. We propose that these two token operations are functionally opposed yet conceptually unified.


\subsection{Training}

\textbf{Trainable Modules.} The encoders used for multimodal representations for audio, video, and text are fully frozen, perceiving the pre-trained knowledge. However, the multimodal fusion transformer and $[seg]$ token are fully trainable during our training process, allowing them to learn how to generate the fused and aligned embeddings tailored for prompting SAM2. For SAM2, we keep the image encoder and memory attention modules frozen while training the prompt encoder and mask decoder. Additionally, all the MLPs for projecting are trainable.

\textbf{Training Strategy.} Our training strategy initially activates SAM2's multimodal prompting capability at the image level and then leverages its extensive pre-trained knowledge for zero-shot video segmentation. In this setting, the fusion transformer and $[seg]$ token are trained with full access to the entire video sequence, while SAM2 is exposed only to the first frame during training. Specifically, we consolidate multimodal information (video, audio, and text) through the fusion module, which compresses these features into tokens that serve as prompts for SAM2 on the first frame, shown in Figure \ref{pipleline} (Training). Then, a supervised loss is applied to this frame to train SAM2’s referring segmentation abilities. During testing, both the fusion module and SAM2 receive the entire video, allowing for comprehensive segmentation across all frames. This approach enables SAM2 to effectively utilize multimodal prompts on the initial frame and then generalize to the complete video sequence through its zero-shot Video Object Segmentation (VOS) capabilities, shown in Figure \ref{pipleline} (Inference).

\textbf{Loss Function.} The model is trained end-to-end under the supervised mask loss $\mathcal{L}_{mask}$, to encourage the SAM2 to predict the correct localized masks. To compute this loss, we combine the per-pixel binary cross-entropy (BCE) loss and DICE loss with their corresponding weights $\lambda_{bce}$ and $\lambda_{dice}$, which can be formulated by:
\begin{equation}
    {\mathcal{L}_{mask}} = {\lambda _{bce}}\textbf{BCE}(\hat M,M) + {\lambda _{dice}}\textbf{DICE}(\hat M,M)
\end{equation}
where $\hat M$ is a single mask prediction for the first frame and $M$ is the label of the first frame for supervision.

\section{Experiments}
\label{sec:method}

\begin{table*}[t]
    \centering
    \scalebox{1}{
    \begin{tabular}{lccccccccccccc}
        \toprule
        \multirow{3}{*}{Method} & \multicolumn{3}{c}{Seen (\%)} & \multicolumn{3}{c}{Unseen (\%)} & \multicolumn{3}{c}{Mix (S+U)} & \multicolumn{1}{c}{NULL} & \multirow{3}{*}{VENUE} \\
        \cmidrule(lr){2-4} \cmidrule(lr){5-7} \cmidrule(lr){8-10} \cmidrule(lr){11-11}
         & $\mathcal{J}$ & $\mathcal{F}$ & $\mathcal{J\&F}$ & $\mathcal{J}$ & $\mathcal{F}$ & $\mathcal{J\&F}$ & $\mathcal{J}$ & $\mathcal{F}$ & $\mathcal{J\&F}$ & s & \\
        \midrule
        AVSBench~\cite{zhou2022avs} & 21.2 & 45.7 & 33.5 & 27.7 & 50.9 & 39.3 & 24.5 & 48.3 & 36.4 & - & ECCV'2022 \\
        \quad \quad \quad \quad \textcolor{blue}{\textit{+text}} & 23.2 & 51.1 & 37.2 & 32.4 & 54.7 & 43.5 & 27.8 & 52.9 & 40.3 & 0.208 & \\
        AVSegFormer~\cite{gao2024avsegformer} & 29.2 & 42.3 & 35.7 & 32.7 & 45.1 & 38.9 & 30.9 & 43.7 & 37.3 & - & AAAI'2024 \\
        \quad \quad \quad \quad\textcolor{blue}{\textit{+text}} & 33.5 & 47.0 & 40.2 & 36.1 & 50.1 & 43.1 & 34.8 & 48.6 & 41.7 & 0.171 & \\
        GAVS~\cite{wang2023prompting}  & 24.8 & 45.5 & 35.2 & 27.8 & 46.0 & 36.9 & 26.3 & 45.7 & 36.0 & - & AAAI'2024 \\
        \quad \quad \quad \quad\textcolor{blue}{\textit{+text}} & 28.9 & 49.8 & 39.4 & 29.8 & 49.7 & 39.8 & 29.4 & 49.8 & 39.6 & 0.190 & \\
        ReferFormer~\cite{Wu_2022_CVPR} & 27.4 & 47.7 & 37.5 & 24.9 & 50.4 & 37.6 & 26.2 & 49.1 & 37.6 & - & CVPR'2022 \\
        \quad \quad \quad \quad\textcolor{blue}{\textit{+audio}} & 31.3 & 50.1 & 40.7 & 30.4 & 48.8 & 39.6 & 30.9 & 49.5 & 40.2 & 0.176 & \\
        R2VOS~\cite{li2023robust} & 21.3 & 30.9 & 26.1 & 23.9 & 43.1 & 33.5 & 22.6 & 37.0 & 29.8 & - & ICCV'2023 \\
        \quad \quad \quad \quad\textcolor{blue}{\textit{+audio}} & 25.0 & 41.0 & 33.0 & 27.9 & 49.8 & 38.9 & 26.5 & 45.4 & 35.9 & 0.183 & \\
        EEMC~\cite{wang2024refavs} & 34.2 & 51.3 & 42.8 & 49.5 & 64.8 & 57.2 & 41.9 & 58.1 & 50.0 & \textbf{0.007} & ECCV'2024 \\
        \hline
        \rowcolor{gray!20} SAM2-LOVE (ours) & \textbf{43.5} & \textbf{51.9} & \textbf{47.7} & \textbf{66.5} & \textbf{72.3} & \textbf{69.4} & \textbf{55.0} & \textbf{62.1} & \textbf{58.5} & 0.23  & CVPR2025 \\
        \rowcolor{gray!20} \textit{vs. prev. SOTA} & \textcolor{red!80!black}{\textbf{+9.3}} & \textcolor{red!80!black}{\textbf{+0.6}} & \textcolor{red!80!black}{\textbf{+4.9}} & \textcolor{red!80!black}{\textbf{+17.0}} & \textcolor{red!80!black}{\textbf{+7.5}} & \textcolor{red!80!black}{\textbf{+12.2}} & \textcolor{red!80!black}{\textbf{+13.1}} & \textcolor{red!80!black}{\textbf{+4.0}} & \textcolor{red!80!black}{\textbf{+8.5}} & - & - \\
        \bottomrule
    \end{tabular}
    }
    \caption{Performance comparison across different methods in Seen, Unseen, Mix (S+U), and NULL settings of Ref-AVS benchmark. The mix indicates the average value of seen and unseen splits.}
    \label{tab1}
\end{table*}

\subsection{Experimental Settings}
\textbf{Dataset.} We evaluate our method on the Ref-AVS benchmark \cite{wang2024refavs}, a substantial collection of 4,000 videos with manual pixel-level annotations and 20,000 expressions reflecting the scenarios in audio, vision, and time dimensions. Additionally, the collected videos contain abundant and diverse visual elements. Each video encompasses a high average number of objects about 1.72, a wide variety of audible objects in 48 categories, and 3 kinds of static objects. Therefore, the tested scenarios are more challenging with multiple sound sources and multiple semantics. The dataset is divided into three splits, a training set of 2908 videos, a validation set of 276 videos, and a test set of 818 videos. The test set is further divided into three distinct subsets: 1) \textit{seen splits} that are composed of the same 39 categories in the training set. 2) \textit{unseen splits} with 13 additional categories that do not appear in the training stage. 3) \textit{null splits} to test the model when an expression refers to an object that either does not exist or is not visible in the given context.

\textbf{Baselines.} We compare our method with six
challenging baselines: AVSBench~\cite{zhou2022avs}, AVSegFormer~\cite{gao2024avsegformer}, GAVS~\cite{wang2023prompting}, ReferFormer~\cite{Wu_2022_CVPR}, R2VOS~\cite{li2023robust} and EEMC~\cite{wang2024refavs}. AVSBench, AVSegFormer, and GAVS are designed for Audio-Visual Segmentation (AVS) tasks while ReferFormer and R2VOS claim for Referring Video Object Segmentation(Ref-VOS) tasks. We also report the results achieved by offering the third their absent modalities, which is consistent with EEMC. EEMC is specially designed to achieve fine-grained semantic understanding in the LAVS and also proposes the Ref-AVS task.


\textbf{Implementation Details.} For the training recipe, we adopt the DeepSpeed training engine with ZeRO-2 to train our model on 2 NVIDIA A100 GPUs.  The AdamW \cite{loshchilov2018decoupled} optimizer is harnessed with the learning rate and weight decay set to 1e-4 and 0, respectively. We also adopt WarmupDecayLR as the learning
rate scheduler, where the warmup iterations are set to 100. The weights of the BCE loss $\lambda_{bce}$ and Dice loss $\lambda_{dice}$ are set to 1.0 and 1.0, respectively. The batch size is set to 8 and the gradient accumulation step is set to 2. For the network architecture, the fusion transformer is the stacked 6 transformer blocks with bi-directional self-attention. The hidden dimension of the transformer is 1024 and the number of $[seg]$ tokens is set to 1. SAM2 is initialized with the weight of SAM2-Hiera-large.

\textbf{Metrics.} Following \cite{wang2024refavs, luo2024soc}, we assess the performance of the Ref-AVS benchmark using the Jaccard index ($\mathcal{J}$), the F-score ($\mathcal{F}$), and their average value ($\mathcal{J\&F}$) as key evaluation metrics.

\subsection{Main Results}
Table \ref{tab1} shows the comparative results of our method on Ref-AVS benchmarks. Our SAM2-LOVE achieves state-of-the-art (SOTA) performance and significantly outperforms the second-best method EEMC on both seen and unseen test splits. The results indicate that our model has highly generalized capabilities that can be directly used for out-of-domain scenarios. We attribute two reasons leading to these huge performance improvements. Firstly, the fusion transformer can effectively step-by-step compress the multimodal cues into the learnable tokens through the $[seg]$ token propagation strategy and $[cls]$ token accumulation. The generated token contains high semantics as well as positional information, which can accurately prompt SAM2 to segment the inserted targets. Secondly, SAM2 showcases strong zero-shot VOS capabilities, accurately tracking the referred object despite the circumstance where it suddenly disappears and occurs again or similar objects have interacted together.

\begin{figure*}[t]
	\centering	\includegraphics[width=\textwidth]{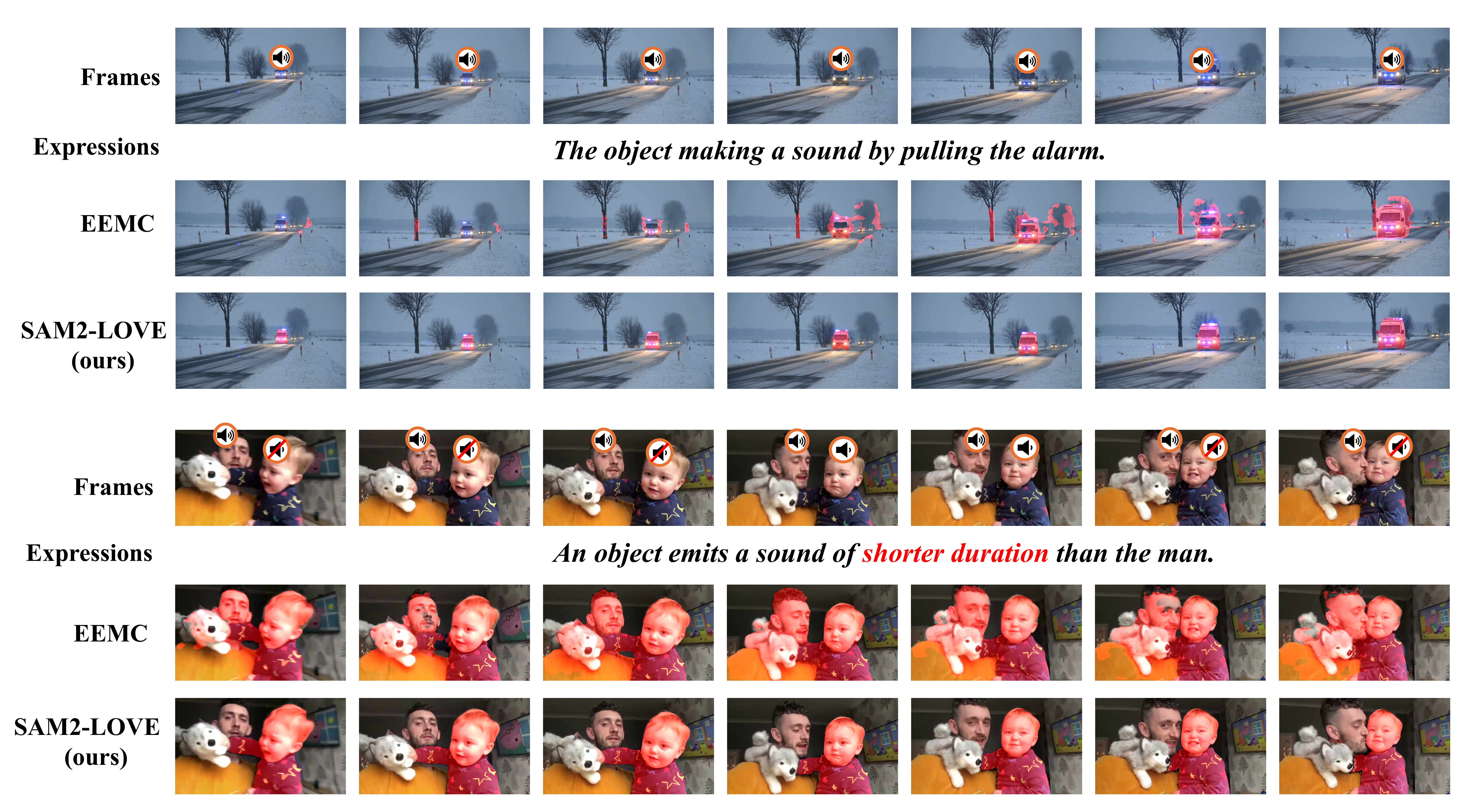}
	\caption{The visualization results of the referred objects in the Ref-AVS. Notably, we use the legends of the trumpet to represent the different sound volumes of the audio objects, from silent to increasing loudness. More cases can be found in the supplementary.}
	\label{comp_vis}
\end{figure*} 

\subsection{Quality Assessment}
Figure \ref{comp_vis} shows the visualization comparative results with the previous SOTA EEMC. In the first case, we can observe that the SAM2 shows strong consistency within different frames, which can track the referred object with high mask quality. Additionally, it also demonstrates that the $[seg]$ token can accurately locate the small referring target, while EEMC fails in the first several frames. The second case shows that our SAM2-LOVE is more robust regarding changes in audible objects. EEMC may not distinguish the difference or malfunction in understanding the hidden meaning in the expressions (i.e. ``shorter duration" instead of the obvious ``the man").

\begin{table}[t]
    \centering
    \scalebox{0.73}{
    \begin{tabular}{ccccccccccc}
        \toprule
        \multirow{2}{*}{\shortstack{CLIP-T \\ CLIP-V}} & \multirow{2}{*}{\shortstack{RoBERTa \\ ViT}} & \multirow{2}{*}{$[cls]$ Acc} & \multicolumn{3}{c}{Seen (\%)} & \multicolumn{3}{c}{Unseen (\%)} \\ 
        \cmidrule(lr){4-6} \cmidrule(lr){7-9}
         &  &  & $\mathcal{J}$ & $\mathcal{F}$ & $\mathcal{J\&F}$ & $\mathcal{J}$ & $\mathcal{F}$ & $\mathcal{J\&F}$ \\
        \midrule
        $\checkmark$ &  & $\checkmark$ & 42.5 & 51.2 & 46.8 & 65.1 & 71.4 & 68.2 \\
         & $\checkmark$ & $\checkmark$ & \textbf{43.5} & \textbf{51.9} & \textbf{47.7} & \textbf{66.5} & \textbf{72.3} & \textbf{69.4} \\
         & $\checkmark$ &  & 42.6 & 50.9 & 46.7 & 66.3 &71.6 & 69.0 \\
        \bottomrule
    \end{tabular}
    }
    \caption{Ablations on backbone design and our proposed $[cls]$ token accumulation strategy.}
    \label{tab2}
\end{table}

\subsection{Ablation Study}
\textbf{Effect of Backbone.} Table \ref{tab2} shows the ablation study of backbones. Compared to CLIP vision and text encoder, RoBERTa and ViT achieve better performance, which can be the difference from the pre-trained paradigm. To align with visual elements, the CLIP textual encoder focuses more on the text with specific semantics such as cat and person. However, in the LAVS scenes, the massive expressions tend to provide a kind of control signal without referred category names, e.g., `` the one making the sound.'' and `` the one making the most intense sound.'' Therefore, the RoBERTa may demonstrate greater strength in understanding instructions due to expertise in natural language understanding during training.

\begin{table}[t]
    \centering
    \scalebox{0.73}{
    \begin{tabular}{cccccccccc}
        \toprule
        \multirow{2}{*}{\shortstack{Number of \\ transformer blocks}} & \multicolumn{3}{c}{Seen (\%)} & \multicolumn{3}{c}{Unseen (\%)} \\
        \cmidrule(lr){2-4} \cmidrule(lr){5-7}
         & $\mathcal{J}$ & $\mathcal{F}$ & $\mathcal{J\&F}$ & $\mathcal{J}$ & $\mathcal{F}$ & $\mathcal{J\&F}$ \\
        \midrule
        1 layer  & 41.5 & 49.4 & 45.4 & 65.2 & 70.9 & 68.1 \\
        6 layers  & \textbf{42.6} & \textbf{50.9} & \textbf{46.7} & 66.3 & 71.6 & 69.0 \\
        12 layers & 42.0 & 50.7 & 46.4 & \textbf{67.7} &\textbf{ 73.3 }& \textbf{70.5 }\\
        \bottomrule
    \end{tabular}
    }
    \caption{Ablations on the number of fusion transformer blocks.}
    \label{tab3}
\end{table}

\textbf{Effect of $[cls]$ Token Accumulation.} Since we harness the propagated token to prompt the first frame for SAM2, the attention for the first several frames may weaken during propagation, causing the token to depend only on the last frame. Therefore, in the fusion transformer, we attach the visual global $[cls]$ tokens of previous frames to the current frame to update the history token, avoiding the $[seg]$ token forgetting the information of former frames. Compared to the last row and second row in Table \ref{tab2}, we can observe that this strategy can effectively improve performance for free without additional design or resource consumption.

\textbf{Effect of Fusion Transformer Design.} The design of our fusion transformer is a naive transformer architecture without additional operations such as ROPE~\cite{su2024roformer} or attention enhancement~\cite{shen2021efficient}. Table \ref{tab3} shows the different performances under different numbers of transformer blocks. Increasing the model's volume can lead to gains in the unseen test set and comparable performance in the seen test set. The single layer can also achieve satisfying results, demonstrating that the simple multimodal fusion still makes a huge impact. However, considering the additional resource consumption and balance of efficiency, we use 6 layers of transformer instead of 12.

\begin{table}[t]
    \centering
    \scalebox{0.88}{
    \begin{tabular}{cccccccccc}
        \toprule
        \multirow{2}{*}{\shortstack{Number of \\ $[seg]$ tokens}} & \multicolumn{3}{c}{Seen (\%)} & \multicolumn{3}{c}{Unseen (\%)} \\
        \cmidrule(lr){2-4} \cmidrule(lr){5-7}
         & $\mathcal{J}$ & $\mathcal{F}$ & $\mathcal{J\&F}$ & $\mathcal{J}$ & $\mathcal{F}$ & $\mathcal{J\&F}$ \\
        \midrule
        1 & \textbf{42.6} &\textbf{ 50.9} & \textbf{46.7} & 66.3 & 71.6 & 69.0 \\
        4 & 41.3 & 50.8 & 46.0 & 65.2 & 71.0 & 68.1 \\
        8 & 40.7 & 49.8 & 45.2 & \textbf{67.0} & \textbf{72.2 }& \textbf{69.6} \\
        \bottomrule
    \end{tabular}
    }
    \caption{Ablations on the numbers of $[seg]$ tokens.}
    \label{tab4}
\end{table}

 \begin{table}[t]
    \centering
    \scalebox{0.83}{
    \begin{tabular}{lcccccc}
        \toprule
        \multirow{2}{*}{Fusion Strategy} & \multicolumn{3}{c}{Seen (\%)} & \multicolumn{3}{c}{Unseen (\%)} \\
        \cmidrule(lr){2-4} \cmidrule(lr){5-7}
         & $\mathcal{J}$ & $\mathcal{F}$ & $\mathcal{J\&F}$ & $\mathcal{J}$ & $\mathcal{F}$ & $\mathcal{J\&F}$ \\
        \midrule
        learnable token & \textbf{42.6} & \textbf{50.9} & \textbf{46.7} & \textbf{66.3} & \textbf{71.6} & \textbf{69.0} \\
        mean   & 41.6 & 50.1 & 45.8 & 65.9 & 71.6 & 68.7 \\
        \bottomrule
    \end{tabular}
    }
    \caption{Ablations on the different fusion strategies for generating the prompt token for SAM2.}
    \label{tab5}
\end{table}

\textbf{Effect of the Number of $[seg]$ Tokens.} We only use a single token to promote SAM2 for localization of the correct target, but the number of tokens entering the propagation process can be multiple. As shown in Table \ref{tab4}, increasing the number of tokens can improve significant performances in the unseen split, while this sacrifices the performances of the seen category. 
We think that it fits the insight from LIBERO~\cite{liu2024libero}. The additional tokens can be regarded as kinds of registers that store additional global information, and this prior knowledge would facilitate learning on new tasks (Unseen tasks). On the other hand, utilizing a single token implicitly encodes global information into that token, which can enhance performance on prior tasks (Seen tasks) but could negatively impact performance on new tasks (Unseen tasks). Despite the representation learned by a single token exhibiting a stronger inductive bias, it is still useful and we set it by default.


\textbf{Effect of Fusion Strategy.} The token used to prompt the SAM2 can be generated by different strategies. Besides using a learnable token, we can directly take the mean of whole the sequence that is generated by concatenating all the frames' embeddings with those of audio and text. As shown in Table \ref{tab5}, our learnable token can behave better since it can aggregate information from different frames gradually. In contrast, the direct mean operation may lose a large amount of inter-frame information. 

\textbf{Effect of Training Modules in SAM2.} To preserve the strong VOS capabilities of SAM2, we freeze the memory attention modules but train the prompt encoder and mask decoder to further adapt the SAM2 to the multimodal clue reference, which can achieve the best results shown in Table \ref{tab6}. Without training the mask decoder, it will cause a huge performance drop due to the low capability to fit the multimodal representation of the $[seg]$ token.

\begin{table}[t]
    \centering
    \begin{tabular}{ccccccc}
        \toprule
        \multirow{2}{*}{\shortstack{Mask \\ decoder}} & \multirow{2}{*}{\shortstack{Prompt \\ encoder}} & \multicolumn{3}{c}{Seen (\%)} \\ 
        \cmidrule(lr){3-5}
         &  & $\mathcal{J}$ & $\mathcal{F}$ & $\mathcal{J\& F}$ \\
        \midrule
        * & * & 34.8 & 44.8 & 39.8 \\
        $\checkmark$ & *  & 42.2 & 50.7 & 46.5 \\
        $\checkmark$ & $\checkmark$ & \textbf{42.5} & \textbf{50.9} & \textbf{46.7} \\
        \bottomrule
    \end{tabular}
    \caption{Ablations on the different training modules in SAM2. ``$*$" denotes the frozen module while ``$\checkmark$" makes the module trainable.}
    \label{tab6}
\end{table} 
\subsection{Limitations}
In this section, we discuss the limitations of our method. As shown in Table \ref{tab1} NULL split, our method has a relatively low capability to reject the users' requirements for a null reference. It is the inherent limitation of the prompting paradigm for localization that SAM2 attends to segment something or some irrelevant regions in the image without any meaning. SAM2-LOVE lags behind the EEMC in the null test with 0.2 in $S$ score, achieving 0.23 similar results with the \textit{AVSBench plus text} method. The $S$ denotes the square root of the ratio of the number of foreground pixels over the background pixels. This can be our future work on how to solve the null reference problem.


\section{Conclusion}
This paper introduces SAM2-LOVE, a novel framework for segmenting reference objects in Language-aided Audio-Visual Scenes, overcoming the limitations of previous dual-modality SAM approaches in handling scenarios that integrate three modalities. We propose a multimodal fusion module that compresses information from all modalities into a single token, which then prompts SAM2 for video segmentation. This compressed $[seg]$ token propagates forward through the video to capture both the spatial and temporal features, while also referring to the accumulated historical tokens backward preserving information from previous frames. Extensive experiments demonstrate the superiority of our model on the Ref-AVS benchmark and highlight the efficiency of our design. Future work will focus on addressing scenarios where the three modality references are misaligned, such as the null reference problem.

\section*{Acknowledgement}
This work was supported in part by National Natural Science Foundation of China (Grant No. 62206153), and in part by Guangdong Natural Science Funds for Distinguished Young Scholar (No. 2025B1515020012).

{
    \small
    \bibliographystyle{ieeenat_fullname}
    \bibliography{main}
}


\end{document}